%% file: main.tex
\def\year{2021}\relax
\documentclass[letterpaper]{article} 
\usepackage{aaai21}  
\usepackage{times}  
\usepackage{helvet} 
\usepackage{courier}  
\usepackage[hyphens]{url}  
\usepackage{graphicx} 
\usepackage{natbib}  
\usepackage{caption} 
\usepackage[switch]{lineno} %
\newcommand*\samethanks[1][\value{footnote}]{\footnotemark[#1]}
\title{Relation-aware Graph Attention Model With Adaptive Self-adversarial Training}
\author{

    Xiao Qin\textsuperscript{\rm 1}\thanks{authors contributed equally.}, Nasrullah Sheikh\textsuperscript{\rm 1}\samethanks, Berthold Reinwald\textsuperscript{\rm 1}, Lingfei Wu\textsuperscript{\rm 2}\\
}

\affiliations{
    \textsuperscript{\rm 1}IBM Almaden Research Center, 
    \textsuperscript{\rm 2}IBM Thomas J. Watson Research Center\\
    \{xiao.qin, nasrullah.sheikh\}@ibm.com, \{reinwald, wuli\}@us.ibm.com

}

\usepackage{graphicx}

\newcommand{\system}{{RelGNN}\xspace}
\newcommand{\q}{\scriptsize\textcolor{black}}

\usepackage{array}
\newcommand{\PreserveBackslash}[1]{\let\temp=\\#1\let\\=\temp}
\newcolumntype{C}[1]{>{\PreserveBackslash\centering}p{#1}}
\newcolumntype{R}[1]{>{\PreserveBackslash\raggedleft}p{#1}}
\newcolumntype{L}[1]{>{\PreserveBackslash\raggedright}p{#1}}

\usepackage{hyperref}
\usepackage{booktabs} 
\usepackage{alltt} 
\usepackage{multirow} 
\usepackage{amsmath}
\DeclareMathOperator*{\argmax}{argmax} 
\DeclareMathOperator*{\argmin}{argmin} 

\usepackage{amsfonts}
\usepackage{graphicx}
\usepackage{enumitem}
\usepackage{algorithm}
\usepackage[noend]{algpseudocode}
\usepackage[most]{tcolorbox}
\usepackage{xspace}
\usepackage{subcaption}

\begin{document}



\maketitle
%

\begin{abstract}

This paper describes an end-to-end solution for the relationship prediction task in heterogeneous, multi-relational graphs. We particularly address two building blocks in the pipeline, namely heterogeneous graph representation learning and negative sampling. 
Existing message passing-based graph neural networks use edges either for graph traversal and/or selection of message encoding functions. Ignoring the edge semantics could have severe repercussions on the quality of embeddings, especially when dealing with two nodes having multiple relations.
Furthermore, the expressivity of the learned representation depends on the quality of negative samples used during training. Although existing hard negative sampling techniques can identify challenging negative relationships for optimization, new techniques are required to control false negatives during training as false negatives could corrupt the learning process. To address these issues, first, we propose \system -- a message passing-based heterogeneous graph attention model. In particular, \system generates the states of different relations and leverages them along with the node states to weigh the messages. \system also adopts a self-attention mechanism to balance the importance of attribute features and topological features for generating the final entity embeddings. Second, we introduce a parameter free negative sampling technique -- adaptive self-adversarial (ASA) negative sampling. ASA reduces the false negative rate by leveraging positive relationships to effectively guide the identification of true negative samples. Our experimental evaluation demonstrates that \system optimized by ASA for relationship prediction improves state-of-the-art performance across established benchmarks as well as on a real industrial dataset.

\end{abstract}

\input{intro.tex}

\input{pre.tex}

\input{method.tex}

\input{exp.tex}

\section{Conclusion}

In this paper, we describe an end-to-end solution for relationship prediction in heterogeneous graphs. Particularly, we propose a message passing-based graph attention model -- \system to exploit the heterogeneity of the neighborhood by leveraging the edge semantics. 
We also introduce an effective hard negative sampling method called ASA which reduces the false negative rate by consulting the positive relationships.
We demonstrate the strength of \system together with ASA by comparing against state-of-the-art methods on relationship prediction across established benchmarks and a real industrial dataset.

\section{Ethical Impact}

The views and conclusions expressed in this section are those of the authors and should not be interpreted as necessarily representing the official policies or endorsements, either expressed or implied, of the affiliated corporation.

The results in this paper are a set of empirically verified techniques which can improve the expressive power of the graph neural networks particularly for processing multi-relational, heterogeneous graphs. In this work, we propose two contributions -- one towards information propagation via attention, second for model training to achieve better generalization.  Our proposed model can leverage edge semantics to determine the importance of information propagated from neighbors to improve the quality of the node embedding. For optimization, we propose the adaptive self-adversarial method which could generate hard negative samples while reducing the false negative rate. We specifically deal with the relationship prediction task which is of particular interest to many application areas ranging from social media, user recommendations, network analysis, and scientific discovery. Our work was initiated with the intention to enable and/or advance capabilities and/or services for our partners and clients whose businesses and practices are legislated by respective laws and policies. The potential applications include fraud detection for finance institutions, product recommendation for large enterprises etc.

While aiming for positive impact such as saving costs and providing better customer services and experiences, and deriving business insights,  we do acknowledge the possibility of negative effects especially when our work is made available to the general public. For example, our proposed techniques could assist scammers to identify more vulnerable subjects in a network of people. Enterprises who commit to not use sensitive data such as certain demographic information for business purposes may indirectly make use of those information derived from the insensitive ones by our algorithm. The other application scenario is online social networks, the possible negative impact is privacy concerns.  Our method may derive information which is otherwise unavailable or hidden such as predicting likes/dislikes of a user, deriving insights on non-obvious relationships between users. However, to the best of our knowledge, our work does not create a new line of threats to the society and the above examples which are open questions that remain true to many other works as well. While it is out of scope of this paper to address such issues, we appreciate and agree with the AAAI 2021 organizing committee for bringing attention to the topic of impact beyond technical contributions.


\bibstyle{aaai21}
\bibliography{autodata}


\end{document}

%% file: intro.tex
\section{Introduction} 

Modern data are often relational in nature with heterogeneous interactions between entities and complicated dependency structures. Knowledge graphs~\cite{kgsurvey} (KGs), as a prime example of heterogeneous graphs, model real-world objects, events or concepts as well as various relations among them. 
Making sense of data in such form has gained tremendous attention in large enterprises as graph insights enable new capabilities or services. In particular, this paper focuses on the task of relationship prediction, i.e. predicting the types of relations between entities with rich attributes where the entities are nodes in a graph and the existing relationships form the edges between them. We propose two building blocks in an end-to-end solution for this task, namely \textit{heterogeneous graph representation learning with relation-aware attention} and \textit{adaptive self-adversarial negative sampling} for model training.

First, while knowledge graph embedding methods~\cite{kgsurvey,distmult,rotate,complex,conve} have achieved great success for relationship prediction for heterogeneous graphs, there has been an increasing interest in leveraging Graph Neural Networks (GNNs) for this task~\cite{rgcn}. While early research on GNNs mainly focuses on homogeneous graphs~\cite{gcn,sage,gnnsurvey,chen2020iterative,chen2019reinforcement}, more recent variations and extensions~\cite{rgcn,han,gatne,metapath2vec,hetgnn,hgt,chen2020toward} cope with the heterogeneity aspect of the nodes and edges. The main idea of using GNNs for heterogeneous graphs is to learn dedicated, type-aware encoding functions for information propagation and aggregation~\cite{hgraphsurvey}. However, existing research tends to ignore the semantics of the edges, that is the edge information is only used either for graph traversal (e.g., meta-path based neighborhood sampling~\cite{metapath2vec}) and/or selection of encoding functions~\cite{rgcn}. Ignoring the semantics of edges during aggregation decreases the expressivity of the node embeddings 
as messages propagated even from the same source may contribute quite differently to its neighbor's embedding.
For example, a buyer entity can have multiple relations with a product such as ``purchase", ``has maintenance contract", and ``has warranty". Besides the difference of messages sent by this product via multiple relations, these messages may not be equally important to characterize the buyer. Existing GNNs, message aggregation methods in particular, tend to ignore such phenomenon missing out on the opportunity to obtain more expressive node embeddings. The key challenge is to design a mechanism for the aggregation function to generate differentiable signals from both heterogeneous neighbors and edges. 

Second, most graph representation learning methods can be unified within a \textit{sampled noise contrastive estimation} framework~\cite{samplednce}, especially for link/relationship prediction task. Random negative sampling has been widely adopted for its simplicity and efficiency. However, it suffers seriously from the \textit{vanishing gradient problem} as most negative samples can be easily classified~\cite{nscaching}. Generative adversarial network (GAN) based solutions~\cite{kbgan,igan} are later proposed for addressing this issue. Using GAN, the generator acts like a negative sampler to identify challenging relationships for the discriminator to learn. However, GANs have more parameters and are generally difficult to train. 

Recently, a line of research exploits the model itself~\cite{rotate,nscaching} to identify the challenging negative samples. These models work under the assumption that the hardness of a negative sample is proportional to the error it causes. Despite their superior performance, they still struggle with the problem of over-training on \textit{false negatives} as the training progresses. Thus, the negative samples that cause extremely high errors would not be trustworthy especially in missing link/relationship prediction scenarios.
How to balance the exploitation of hard negative samples, and avoiding false negatives becomes a challenging problem since a clear cut boundary between them does not exist. Moreover, the decision boundaries could vary from case to case due to the heterogeneity of the graphs failing the simple uniform thresholding approaches~\cite{pinsage,rotate,nscaching}.


\noindent\textbf{Our Approach}. To address the above issues, we propose \system -- a graph attention model for learning entity embeddings in heterogeneous graphs and a negative sampling algorithm called adaptive self-adversarial (ASA) negative sampler for model training.

\system follows the \textit{spatial approach}~\cite{gnnsurvey} based on the \textit{message passing paradigm}~\cite{mpgoogle} in generating embeddings of entities in an \textit{inductive} manner~\cite{sage}. \system starts with the attribute embedding layers configured for different node type.  It consists of multiple graph convolution layers and the messages collected from the neighbors are encoded differently according to their edge type. 
Besides, the messages are weighed via a multi-head attention mechanism which considers the states of the nodes on both ends of the edge as well as the state of the edge since two entities could have multiple relations in the heterogeneous graph. To balance the importance of the attribute features and the topological features for the relationship prediction, the final embedding combines the attribute embedding and the output of the message propagation using a self-attention mechanism. 


To capture the dynamic distribution of the negative samples during training, ASA algorithm makes use of the trained model from the prior iteration to evaluate the hardness of a generated negative sample, i.e. the higher the gradient it causes, the harder the sample is to the model. To overcome the false negative issue, ASA also evaluates and considers the hardness of the positive samples. The idea, inspired by a negative sampling method for training dialogue systems~\cite{adaptivens}, is based on the following heuristic: since the negative samples are derived from an existing relationship, the level of confidence of the model on the positive sample should match its level of confidence on the negative one. If the gradient caused by a negative sample is way higher, that particular sample could be a false negative, and therefore should be avoided.

It is worthwhile to highlight the following contributions of this paper:
\begin{itemize}[leftmargin=*]

\item We propose \system -- a message passing-based graph attention model for heterogeneous graph which leverages the edge semantics to boost the expressive power.

\item We propose ASA -- a parameter free negative sampling technique that identifies hard negative samples while effectively reducing the false negative rate.

\item We evaluate \system and ASA negative sampler by an ablation study and a position study which compares our method against state-of-the-art approaches in relationship prediction task on inductive benchmarks as well as a real industrial dataset.


\end{itemize}

%% file: pre.tex
\section{Preliminary}

In this section, we define the heterogeneous graph and introduce the main notations. Then, we introduce the message passing framework~\cite{mpgoogle} upon which we build \system.

\begin{figure*}[t]
\centering
\includegraphics[width=0.85\textwidth]{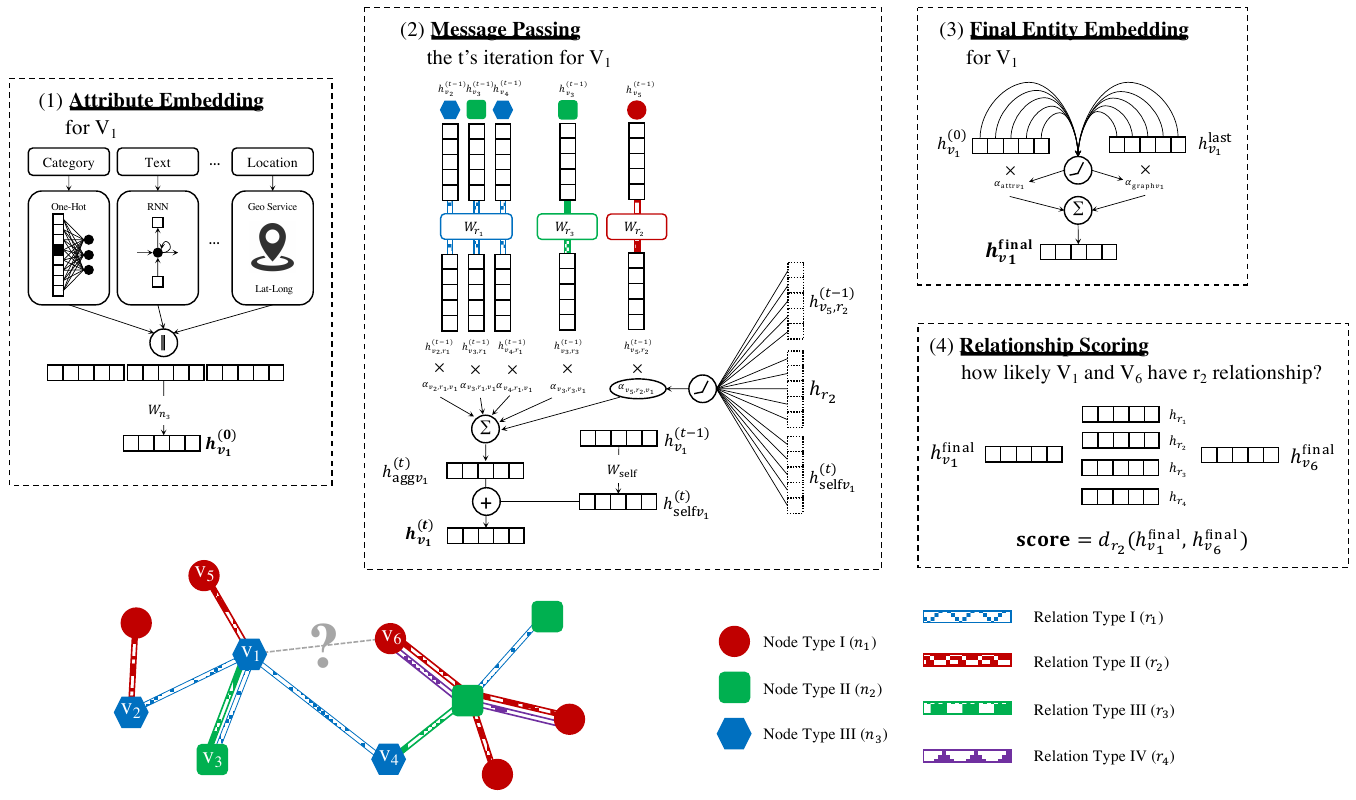}
\caption{Illustration of \system. (1) an attribute embedding network for a specific node type. (2) a message propagation process for $v_1$ at time $t$. (3) the self-attention mechanism for generating the final embedding of $v_1$. (4) relationship scoring.}
\label{fig:overview}
\end{figure*}

\subsection{Attributed Heterogeneous Graph}
Let $G = (V,E,A,R,\phi)$ denote a heterogeneous graph. $V = \{v_1,\cdots,v_n\}$ is a set of nodes in $G$. $A = \{a_1,\cdots,a_k\}$ defines the node attribute schema of $G$. Each node $v_i \in V$ is associated with a node type $\phi(v_i)$. Its corresponding attribute schema is a subschema of $A$ denoted by $A_{\phi(v_i)} \subseteq A$. $E = \{e_1,\cdots,e_m\}$ is a set of directed edges and $R$ defines the edge types in $G$. Each edge $e_k = (v_i, r_i ,v_j)$ indicates a relationship from $v_i$ to $v_j$ of a relation type $r_i \in R$. Note that in this definition, two nodes can have multiple relations. In the rest of the paper, the terms node and entity, edge and relationship are used interchangeably.



\subsection{Message Passing-based GNN}
\label{sec:mp}

A message passing-based GNN~\cite{mpgoogle} adopts a neighborhood aggregation strategy where the representation of a node is learned in an iterative manner by aggregating representations of its neighbors. After $t$ iterations (layers) of aggregation, a node's representation encodes the structural information within its $k$-hop neighborhood. Formally, the $t$-th layer of a GNN is:

\begin{equation}
\label{eq:aggr}
h^{(t)}_{\text{agg}v_i} = Aggregate^{(t)}\big(\big\{h^{(t-1)}_{v_j} | v_j \in Neighbor(v_i)\big\}\big),
\end{equation}

\begin{equation}
\label{eq:comb}
h^{(t)}_{v_i} = Combine^{(t)}\big(h^{(t-1)}_{v_i},h^{(t)}_{\text{agg}v_i}\big),
\end{equation}

\noindent where $h^{(t)}_{v_i}$ is the vector representation of the node $v_i$ at the $t$-th iteration. In inductive learning setting, the node state is usually initialized by the node attributes embedding. The exact definitions of $Neighbor(\cdot)$, $Aggregate^{(t)}(\cdot)$ and $Combine^{(t)}(\cdot)$ in GNNs define their modeling approaches~\cite{gin}.

%% file: method.tex
\section{Methodology}
In this section, we introduce \system -- a message passing-based graph attention neural network for generating entity embeddings for a heterogeneous graph, and a new negative sampling method, called ASA to train \system for relationship prediction.  

As illustrated in Figure~\ref{fig:overview}, \system consists of four components.
The node heterogeneity in the \textbf{attribute embedding} phase is dealt by configuring dedicated attribute embedding networks for each node type and eventually bringing the attribute embeddings of different node types into the same space. An attribute embedding network for a specific node type consists of individual embedding layers or methods appropriate for different modalities and types defined by the node attribute schema. For example, the attribute embedding network for $v_1$ in Figure~\ref{fig:overview}(1) consists of a multilayer perceptron (MLP) network to process categorical attribute in one-hot-encoding, a recurrent neural network-based language model~\cite{rnnlm} to process text attribute and a geo-location API to obtain the latitude and longitude of an address. The outputs are concatenated and transformed by a weight matrix to the final attribute embedding of $v_i$ denoted by $h^{(0)}_{v_i}$.

\system learns topological features of nodes through \textbf{message passing}. The key idea is to encode and weigh the messages differently by fully incorporating edge heterogeneity and node state. We will discuss this in detail in Section~\ref{sec:mp}. \system then generates the final entity embedding by combining the node attribute embedding and node graph embedding via a self-attention mechanism.

For relationship prediction, \system adopts factorization approaches~\cite{distmult,complex} to compute how likely two nodes share a specific relation by considering the embedding of the two nodes as well as the corresponding relation embedding. 
Finally, we introduce a new negative sampling method to efficiently train \system in Section~\ref{sec:ns}. It can be generalized to train a broad spectrum of GNNs for link/relationship prediction task.

\subsection{Attention-based Message Passing Network for Attributed Heterogeneous Graphs}
\label{sec:relgnnmp}



\textbf{Message Propagation}. We start by describing the node embedding generation process in a single message passing iteration. For node $v_i$, we define a propagation function for calculating the forward-pass update in $G$:

\begin{equation}
\begin{aligned}
    h_{v_i}^{(t)} = \sigma\Big(\sum_{r\in R}\sum_{v_j \in N^{r}_{v_i}} \alpha_{(v_i,v_j)} W^{(t-1)}_{r}h_{v_j}^{(t-1)} + W_{self}^{(t-1)}h^{(t-1)}_{v_i}\Big),
\end{aligned}
\label{eq:pass}
\end{equation}

\noindent where $\sigma$ is an activation function, $N^{r}_{v_i}$ denotes the set of immediate neighbors of $v_i$ which connect with $v_i$ via $r \in R$, $W^{(t-1)}_{r}$ is a weight matrix dedicated for encoding messages sent via a particular relation $r$, $W^{(t-1)}_{self}$ is a weight matrix that encode $v_i$'s embedding from the last iteration, and $\alpha_{(v_i,v_j)}$ is an attention weight for the message sent from $v_j$. The first term in Equation~\ref{eq:pass} defines our aggregation function in the message passing network (Equation~\ref{eq:aggr}). First, the messages sent via different types of edges are encoded differently by introducing a dedicated weight matrix~\cite{rgcn}. To avoid overfitting, we apply \textit{basis-decomposition} on these matrices. Second, the messages are weighed by considering the nodes as well as the relation. It addresses the issue of having multiple relations between two nodes, e.g. $v_3$ and $v_1$ in Figure~\ref{fig:overview}, which is typical in heterogeneous graphs. In particular, we use a multi-head attention mechanism to compute the attention coefficient. 
A single head attention weight in Equation~\ref{eq:pass} is expressed as: $\alpha_{(v_i,v_j)} =$

\begin{equation}
\scriptsize
\begin{aligned}
     \frac{\exp{\big(\sigma\big(\boldsymbol{a_e}^{\top}[W^{(t-1)}_{self}h^{(t-1)}_{v_i}\parallel h_{r}\parallel W^{(t-1)}_{r}h^{(t-1)}_{v_j}]\big)\big)}}{\sum_{r'\in R}\sum_{v_n \in N^{r'}_{v_i}}\exp{\big(\sigma\big(\boldsymbol{a_e}^{\top}[W^{(t-1)}_{self}h^{(t-1)}_{v_i}\parallel h_{r'}\parallel W_{r'}^{(t-1)} h^{(t-1)}_{v_n}]\big)\big)}}
\end{aligned}
\label{eq:atten}
\end{equation}

\noindent where $\cdot^{\top}$ represents transposition, $\parallel$ is the concatenation operation,  $h_{r}$ is the state of the relation $r$ and $\boldsymbol{a_e}$ is weight vector for a single head attention mechanism. In our experiment, we use the LeakyReLu as the activation function. The above computation is illustrated in Figure~\ref{fig:overview}(2). The final multi-head attention-based propagation is then expressed as: $h_{v_i}^{(t)} =$

\begin{equation}
\small
\begin{aligned}
     \sigma\Big(\frac{1}{L}\sum^{L}_{l=1}\Big(\sum_{r\in R}\sum_{v_j \in N^{r}_{v_i}} \alpha^{l}_{(v_i,v_j)} W^{(t-1)}_{r}h_{v_j}^{(t-1)} + W_{self}^{(t-1)}h^{(t-1)}_{v_i}\Big)\Big),
\end{aligned}
\label{eq:mp}
\end{equation}

\noindent where $L$ is the number of heads. 

\noindent\textbf{Final Entity Embedding}. Based on our observation, the attribute embedding itself sometimes provides good expressive power to make the relationship prediction, and such cases vary depending upon the particular nodes and edges. Therefore, we use a self-attention mechanism to balance the contribution of the attribute embedding $h^{(0)}_{v_i}$ and the final graph embedding $h^{last}_{v_i}$:

\begin{equation}
\begin{aligned}
    h_{v_i}^{\text{final}} = \alpha_{attr}h^{(0)}_{v_i} + \alpha_{graph}h^{last}_{v_i},
\end{aligned}
\label{eq:finalpass}
\end{equation}

\begin{equation}
\begin{aligned}
     \alpha_{attr} = \frac{\exp(\sigma(\boldsymbol{a_s}^{\top}h^{(0)}_{v_i}))}{\exp(\sigma(\boldsymbol{a_s}^{\top}h^{(0)}_{v_i}))+\exp(\sigma(\boldsymbol{a_s}^{\top}h^{last}_{v_i}))},
\end{aligned}
\label{eq:finalattn}
\end{equation}


\noindent where $\boldsymbol{a}_s$ is a shared attention vector. $\alpha_{graph}$ is computed in the same way as Equation~\ref{eq:finalattn}. Specifically, we use a multi-head self-attention mechanism and employ \textit{averaging} for aggregating states from different heads.

\subsection{Adaptive Self-Adversarial Negative Sampling}
\label{sec:ns}

\textit{Negative sampling}~\cite{word2vec} is a technique to approximate a \textit{softmax} function with a huge output layer. It is based on the idea of noise contrastive estimation which says that a good model should differentiate positive signals from negative ones. It has been adapted to graph learning problems~\cite{samplednce} especially where the learning objective involves predicting relations between nodes:

\begin{equation}
\scriptsize
\begin{aligned}
    \argmin_\theta \sum_{(v_i,r,v_j) \in E} \big[\ell(+1, d_r(f(v_i),f(v_j))) + \ell(-1, d_r(f(\bar{v}_m),f(\bar{v}_n)))\big],
\end{aligned}
\end{equation}

\noindent where $\theta$ denotes the model parameters, $\ell$ is usually defined as cross entropy, $\bar{v}_m,\bar{v}_n \in V$ are parts of a negative relationship sample, i.e. $(\bar{v}_m,r,\bar{v}_n) \notin E$, $f$ refers to a graph embedding network (\system in our case), and $d_r$ is a scoring function. In this paper, we define $d_r$ as DistMult~\cite{distmult} factorization which can be replaced by other \textit{relation learning} models~\cite{complex} as long as it learns embeddings for relations:

\begin{equation}
\begin{aligned}
    d_r(f(v_i),f(v_j)) = softmax(h_{v_i}^{\text{final}\top} M_r h_{v_j}^{\text{final}}),
\end{aligned}
\label{eq:score}
\end{equation}

\noindent where $M_r \in \mathbb{R}^{d \times d}$ is a diagonal matrix associated with $r \in R$. $h_r$ is the diagonal of $M_r$ which is used in Equation~\ref{eq:atten} for computing the attention coefficient.

A negative sample in the context of relationship prediction is normally generated by altering one of the nodes of an existing edge while keeping the relation type fixed, e.g. $(v_i,r,\bar{v}_n) \notin E$ or $(\bar{v}_m,r,v_j) \notin E$. Random negative sampling is commonly adopted due to its simplicity and efficiency. However, \cite{nscaching} has shown that only a small subset of all possible negative samples are useful for training while most of the negative samples are trivial cases from which the model does not gain much discriminative power. 

Recently, GAN-based solutions~\cite{kbgan,igan} have attracted a lot of attention for this particular problem. The idea is to use the generator as a negative sampler to generate hard negative relationships for the discriminator. However, GANs introduce more parameters and are generally difficult to train. To overcome the computational barrier, self-adversarial negative sampling has been proposed~\cite{rotate,nscaching}. The core idea is to use the model itself to evaluate the hardness of negative samples, i.e., when using the \textit{stochastic gradient descent} algorithm for optimization, a negative sample with a high gradient for the current model is a hard negative sample. Let ${d'}_r$ and $f'$ be the scoring and embedding function trained from the previous iteration, then for a relationship $(v_i,r,v_j) \in E$, the negative sample selection is defined as:

\begin{equation}
\begin{aligned}
    \argmax_{\{\bar{v}_m,r,\bar{v}_n\} \notin E} {d'}_r(f'(\bar{v}_m),f'(\bar{v}_n)),
\end{aligned}
\label{eq:sa}
\end{equation}

\noindent meaning that the negative sample which is predicted as a positive relationship with a high score is preferred to be used in the current training process. To reduce the computational cost of evaluating every possible negative sample to optimize Equation~\ref{eq:sa}, NSCache~\cite{nscaching} only evaluates a small pool of negative samples randomly selected from the complete set and samples the ones only from this small pool. Although such methods offer superior performance than GAN-based solutions, they do not solve the problem of introducing false negatives because Equation~\ref{eq:sa} always assigns the highest sampling probability to the worst mispredictions. To tackle this, we propose the adaptive self-adversarial (ASA) negative sampling method. Since a negative sample is a variation of the positive relationship, the idea is to make use of the positive relationship during the evaluation to control the hardness accordingly. Specifically, we alter the selection strategy from Equation~\ref{eq:sa} to:

\begin{equation}
\begin{aligned}
    \argmin_{\{\bar{v}_m,r,\bar{v}_n\} \notin E} |{d'}_r(f'(v_i),f'(v_j)) - {d'}_r(f'(\bar{v}_m),f'(\bar{v}_n)) - \mu|,
\end{aligned}
\label{eq:asa}
\end{equation}

\noindent where $\mu$ is a positive constant. Instead of forcing the selection towards the hardest negative sample, ASA selects a negative sample with moderate difficulty by considering the score on the respective positive sample. That is, the score of a selected negative sample may not be higher than the score of the positive relationship from which it is derived. The implication is that the negative sample with score higher than its respective positive relationship can be a false negative. We also introduce a margin $\mu$ as a hyperparameter to further control the hardness -- the higher the $\mu$ the easier the case. Inspired by the self-paced learning ~\cite{selfpaced}, we also experiment with different decay functions, e.g. exponential and linear decay function for $\mu$ to increase the hardness as the training progresses.

%% file: exp.tex
\section{Experimental Evaluation} 
This section summarizes our experimental setup and reports, and analyzes the measurements. 

\begin{table*}[ht]
\centering
\caption{Statistics of the datasets.}
\begin{tabular}{@{}p{0.22\textwidth}*{5}{R{\dimexpr0.12\textwidth-2\tabcolsep\relax}}@{}}
\toprule
{\bf Dataset} & {\bf Nodes} & {\bf Edges} & {\bf Attr. Dim} & {\bf Relations} & {\bf\scriptsize (train/valid/test)}\\
\midrule
Amazon~\cite{gatne} & 10,166 & 148,865 & 1,156 & 2 &85/5/10\\
Youtube~\cite{gatne} & 2,000 & 1,310,617 & 2,000 & 5&85/5/10\\
Company  & 11,585 & 90,262 & 775 & 6&80/10/10\\
\bottomrule
\end{tabular}
\label{tab:data}
\end{table*}

\noindent{\bf Dataset.} We use two established heterogeneous graph datasets, namely Amazon and Youtube~\cite{gatne} which come with standard validation and test sets.  
To evaluate our method in a real world scenario, we use a proprietary enterprise heterogeneous graph dataset Company concerning buy/sell transactions. 
Company includes challenging validation and test sets by introducing high quality negative samples based on our knowledge of the graph. Basic statistics of these datasets are reported in Table~\ref{tab:data}. 


\noindent{\bf Metric.} We adopt two kinds of evaluation criteria. First, following the common practice, we treat the relationship prediction as a binary classification task, i.e. given two entities and a relation type, predict whether or not such combination holds. We report micro AUC (area under the curve) of ROC (receiver operating characteristic curve) and PRC (precision-recall curve) as well as micro F1 with a fixed cutoff threshold.
Second, to evaluate the negative sampling techniques for model training, we report filtered MRR (mean reciprocal rank) and Hit@k. These ranking metrics evaluate how well the model can separate a positive relationship from all possible negative relationships which are generated by altering one of the entities in the positive relationship. Each test case consists of one positive relationship and all the possible negative samples with the same relation type and at least one of the node in the positive relationship. 
MRR indicates the overall ranking performance of all cases. Hit@k computes the proportion of test cases where the positive relationship appears in the top k results.

\noindent{\bf Baselines.} We compare \system against the state-of-the-art relationship prediction models which can be classified into three categories:

\begin{itemize}[leftmargin=*]
  \item \textbf{Relation learning methods}, widely studied in the context of knowledge graph embedding learn a scoring function which evaluates an arbitrary relationship involving two entities and a specific relation. \textbf{DistMult}~\cite{distmult} exploits a similarity-based scoring function. Each relation is represented by a diagonal matrix and the scoring function is defined as a bilinear function. \textbf{ComplEx}~\cite{complex} extends this idea by introducing the entities and relations into a complex space. \textbf{ConvE}~\cite{conve} utilizes 2-D convolution over embeddings and multiple layers of nonlinear features to model the interactions between entities and relations.
  
  \item \textbf{Proximity-preserving methods} capture the topological information by preserving different types of proximity among the entities in the graph. In particular, we compare against a popular random walk-based method \textbf{metapath2vec}~\cite{metapath2vec}. It utilizes the paths traversed by meta-path guided random walks to model the context of an entity regarding heterogeneous semantics.
  
  \item \textbf{Message passing methods}, as discussed in Section~\ref{sec:mp}, learn the entity embedding by aggregating the neighbors' information. \textbf{R-GCN}~\cite{rgcn} considers the heterogeneity by having a dedicated aggregation function for each edge type. \textbf{HAN}~\cite{han} utilizes meta-path to model higher order proximity. \textbf{GATNE}~\cite{gatne} learns multiple embeddings for an entity, each of which encodes the entity w.r.t a specific edge type.
\end{itemize}

We compare ASA against state-of-the-art negative sampling techniques. Self-adversarial negative sampling strategies~\cite{rotate,nscaching} have shown a superior performance than generative adversarial network (GAN)-based solutions~\cite{kbgan,igan} both in terms of effectiveness and efficiency. We choose the latest variation of the self-adversarial negative sampling method \textbf{NSCaching} as the baseline. It provides some flexibility of avoiding the false negative which will be discussed in Section~\ref{sec:nse}. For simplicity, we refer to the combination of \system and ASA as \system and use $\pm$ASA to denote whether a model is optimized by ASA.

\subsection{Position Study} In this set of experiments, we evaluate and compare \system against state-of-the-art methods.

\noindent{\bf Main Result.}
\begin{table*}[t]
 \centering
\caption{Test results on benchmarks and the real-world dataset. \textbf{AUC} denotes the \textit{area under the receiver operating characteristic curve} value and \textbf{AP} denotes the \textit{average precision} corresponding to the \textit{area under the precision-recall curve} value. $\uparrow$ indicates that the higher the score the better the performance. \textbf{AUC}, \textbf{AP} and \textbf{F1} are reported as percentage. The cutoff threshold for \textbf{F1} is 0.5. {\small\textcolor{black}{($\cdots$)}} after each score indicates the ranking of the method w.r.t. the specific setting. Underlined numbers are quoted from~\cite{gatne}.}
\begin{tabular}{@{}p{0.14\textwidth}*{9}{L{\dimexpr0.095\textwidth-2\tabcolsep\relax}}@{}}
\toprule
& \multicolumn{3}{c}{\bf Amazon} & \multicolumn{3}{c}{\bf Youtube} & \multicolumn{3}{c}{\bf Company} \\
\cmidrule(l){2-4} \cmidrule(l){5-7} \cmidrule(l){8-10}
{\bf Methods} & {\bf AUC$\uparrow$} & {\bf AP$\uparrow$} & {\bf F1$\uparrow$(0.5)}& {\bf AUC$\uparrow$} & {\bf AP$\uparrow$} & {\bf F1$\uparrow$(0.5)}& {\bf AUC$\uparrow$} & {\bf AP$\uparrow$} & {\bf F1$\uparrow$(0.5)}\\
\midrule
ComplEx  & 53.18\q{(10)} & 53.18\q{(10)} & 54.39\q{(10)} &  52.11\q{(10)} & 51.60\q{(10)} & 50.97\q{(9)} & 56.31\q{(8)} & 55.30\q{(9)} & 55.33\q{(8)}\\
ConvE  & 49.65\q{(11)} & 49.79\q{(11)} & 66.38\q{(8)} & 50.03\q{(11)}& 50.07\q{(11)}& 27.32\q{(11)} & 52.43\q{(10)} & 53.17\q{(10)} & 29.15\q{(10)}\\
DistMult & 53.94\q{(9)} & 53.51\q{(9)} & 53.71\q{(11)} & 52.49\q{(9)} & 52.12\q{(9)} & 49.50\q{(10)} & 55.29\q{(9)} & 56.26\q{(8)} & 56.51\q{(7)}\\
DistMult + ASA & 64.85\q{(8)} & 67.29\q{(8)} & 64.30\q{(9)} & 76.45\q{(6)} & 78.63\q{(6)} & 69.48\q{(6)} & 67.86\q{(7)} & 68.72\q{(6)} & 66.47\q{(5)}\\
\midrule
 metapath2vec & \underline{94.15}\q{(6)} & \underline{94.01}\q{(6)} & \underline{87.48}\q{(6)} & \underline{70.98}\q{(7)} & \underline{70.02}\q{(7)} & \underline{65.34}\q{(8)} & 73.47\q{(3)} & 70.88\q{(4)} & 16.35\q{(11)} \\ 
 HAN  & 87.57\q{(7)} & 88.15\q{(7)} & 77.35\q{(7)} & 64.66\q{(8)} & 61.24\q{(8)} & 65.45\q{(7)} & 52.33\q{(11)} & 52.06\q{(11)} & 41.88\q{(9)} \\
 GATNE  & \underline{96.25}\q{(5)} & \underline{94.77}\q{(5)} & \underline{91.36}\q{(4)} & \underline{84.47}\q{(5)} & \underline{82.32}\q{(5)} & \underline{76.83}\q{(5)} & 69.72\q{(5)} & 67.22\q{(7)} & 61.87\q{(6)}\\
 R-GCN & 97.16\q{(4)} & 95.87\q{(4)} & 94.52\q{(2)} & 92.38\q{(4)} & 92.18\q{(4)} & 83.35\q{(3)} & 68.69\q{(6)} & 69.95\q{(5)} & 66.75\q{(4)} \\
 R-GCN + ASA & 98.37\q{(3)} & 97.87\q{(3)} & 94.21\q{(3)} & 93.19\q{(3)} &93.11\q{(3)} & 79.33\q{(4)} & 69.96\q{(4)} & 73.20\q{(3)} & 67.90\q{(3)} \\
\midrule
 RelGNN - ASA & 98.84\q{(2)} & 98.55\q{(2)} & \textbf{95.17}\q{(1)} & 94.39\q{(2)} & 93.41\q{(2)} & \textbf{86.32}\q{(1)} & 74.94\q{(2)} & 73.91\q{(2)} & 70.45\q{(2)}\\
 RelGNN  & \textbf{99.14}\q{(1)} & \textbf{99.01}\q{(1)} & 90.44\q{(5)} & \textbf{96.44}\q{(1)} & \textbf{96.08}\q{(1)} & 83.89\q{(2)} & \textbf{77.68}\q{(1)} & \textbf{77.78}\q{(1)} & \textbf{74.43}\q{(1)} \\
\bottomrule
\end{tabular}
\label{table:benchmarks}
\end{table*}
We first evaluate the performance of \system on the benchmarks. Following the same fashion, we select some negative samples for the validation and test partitions of the Company dataset as well. The task is to predict whether or not a given relationship is true. The overall performance of different methods are reported in Table~\ref{table:benchmarks}. 

Overall, \system (\system trained with ASA negative sampling) performs better than the message passing baselines as well as the proximity preserving baseline. The relation learning methods which do not consider the topological features have lower measurements among all. \system achieves the best AUC and AP on the benchmarks and has the best F1 as well on the Company dataset. We use a fixed threshold 0.5 for computing F1 as it seems natural if one interprets the score as a probability. Although the potential best F1 a model can achieve is reflected by AP, showing the F1 with a uniform threshold demonstrates the difference in score distribution between different models. R-GCN models fall short by relatively small margins especially on Amazon and Youtube due to their capability of modeling the heterogeneous neighborhoods with dedicated message encoding functions. However, R-GCN is not as good on Company due to its lack of emphasis on the multi-relational phenomenon and the attribute information in the final entity embedding. GATNE's performances are consistent across all three datasets. Although GATNE's final embeddings are edge type aware and are mix of topological and attribute information, they do not provide sufficient expressive power to beat \system partially due to their lack of emphasis on the importance of aggregated information via different edge types. The metapath2vec models come after the previous two baselines. However, it gives better performance on the Company dataset especially in terms of AUC and AP. In sum, \system consistently outperforms the competitors and could benefit from ASA training.

\begin{table*}[t]
	\begin{minipage}{0.6\linewidth}
	\caption{Test results on \textit{Company}. \textbf{Hit@k} is reported as a percentage.}
\begin{tabular}{@{}p{0.25\textwidth}*{4}{L{\dimexpr0.175\textwidth-2\tabcolsep\relax}}@{}}
\toprule
& \multicolumn{4}{c}{\bf Company} \\
\cmidrule(l){2-5}
{\bf Methods} & {\bf MRR}$\uparrow$ & {\bf Hit@1}$\uparrow$ & {\bf Hit@10}$\uparrow$ & {\bf Hit@30}$\uparrow$\\
\midrule
Random & .0457\q{(7)} & 2.37\q{(7)} & 7.71\q{(7)} & 14.94\q{(7)}\\
\midrule
NSCaching$_{10}$ & .0789\q{(2)} & \textbf{5.38}\q{(1)} & 11.77\q{(4)} & 19.20\q{(4)}\\
NSCaching$_{100}$& .0695\q{(5)} & 3.89\q{(6)} & 11.50\q{(5)} & 18.09\q{(5)}\\
NSCaching$_{500}$& .0658\q{(6)} & 3.99\q{(4)} & 10.84\q{(6)} & 16.75\q{(6)}\\
\midrule
ASA$_{10}$  & \textbf{.0818}\q{(1)} & 5.18\q{(2)} & 13.32\q{(3)} & 22.01\q{(3)}\\
ASA$_{100}$ & .0754\q{(3)} & 3.95\q{(5)} & \textbf{14.35}\q{(1)} & \textbf{23.71}\q{(1)}\\
ASA$_{500}$ & .0751\q{(4)} & 4.32\q{(3)} & 13.71\q{(2)} & 22.99\q{(2)}\\
\bottomrule
\end{tabular}
\label{tab:company}
	\end{minipage}\hfill
	\begin{minipage}{0.4\linewidth}
		\centering
		\includegraphics[width=\textwidth]{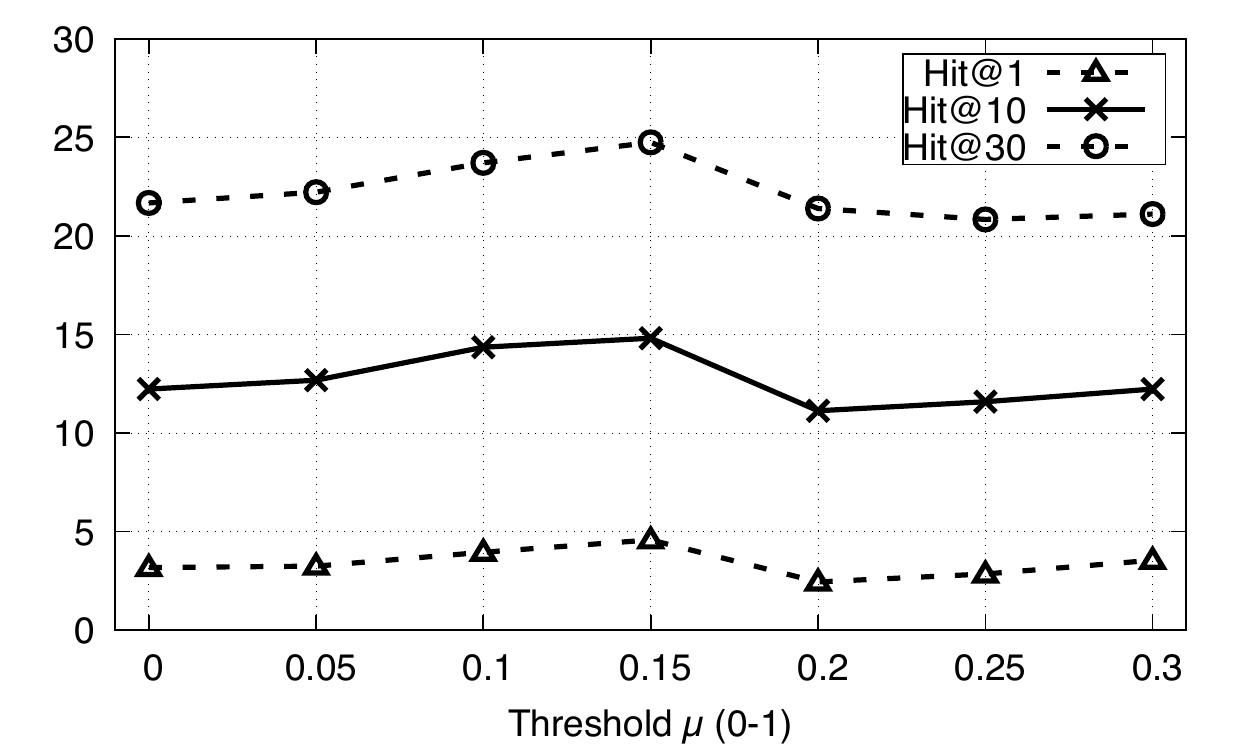}
		\captionof{figure}{Hit@k by varying $\mu$.}
		\label{fig:th}
	\end{minipage}
\end{table*}


\noindent{\bf ASA Negative Sampling.}
\label{sec:nse}
Next, we report the results of a set of experiments which compares our proposed ASA negative sampler against other sampling methods. Since Amazon and YouTube are number coded, we focus on the Company dataset from which we can really tell whether or not a sampled relationship is a false negative. 
We use MRR and Hit@k to evaluate the methods where the perfect scores can be achieved by scoring the true relationships higher than all the respective negative samples.

Table~\ref{tab:company} reports the performance of \system trained by different negative sampling techniques. We vary the pool size for NSCaching and ASA (constant $\mu$ = 0.1). A negative sample pool is randomly sampled from the complete negative sample set. NSCaching and ASA only evaluate and pick the negative samples for training from the pool. \cite{nscaching} points out that a smaller pool size to some degree can help the sampler avoid selecting false negatives. The reason is intuitive which is that a smaller pool will likely to have less amount of false negative than a larger one. The NSCaching and the ASA with pool size as 1 would be equivalent to the Random method. As expected, Random sampler gives the worst results as it fails to identify more difficult ones for the model to learn. Even with a very small pool size, ASA$_{10}$ outperform NSCaching$_{10}$ in most of the measures. As the pool size grows bigger, NSCaching and ASA are able to explore more negative samples. However, without effective mechanism to avoid false negatives, NSCaching's performance suffers as it evaluates more negative samples. Most of the measurements go lower as the pool size increases. On the other hand, ASA sampler can explore more negative samples while keeping lower false negative rate. As a result, ASA samples with larger pool size have better performance especially in Hit@k in this case.

Next, we show how $\mu$ can affect the model performance. We measure the Hit@k on Company of \system trained by ASA samplers  with different $\mu$. As shown in Figure~\ref{fig:th}, we observe that there is an increasing trend of Hit@k up until a specific $\mu$ value and the measurements decrease as the $\mu$ further increases. This demonstrates the effectiveness of $\mu$ on controlling the hardness of the negative samples used for the training. When the $\mu$ is too small, it causes the ASA to select some hard ones which may be false negatives. When the value is too large, most of the negative samples are then trivial cases from which the model may not gain too much. Figure~\ref{fig:th} also shows that a good $\mu$ (0.15) value to have negative samples with a right level of hardness is relatively easy to find during hyperparameter tuning. 

\begin{figure}[t]
	\centering
	\begin{subfigure}{0.45\textwidth}
		\centering
		\includegraphics[width=1\textwidth]{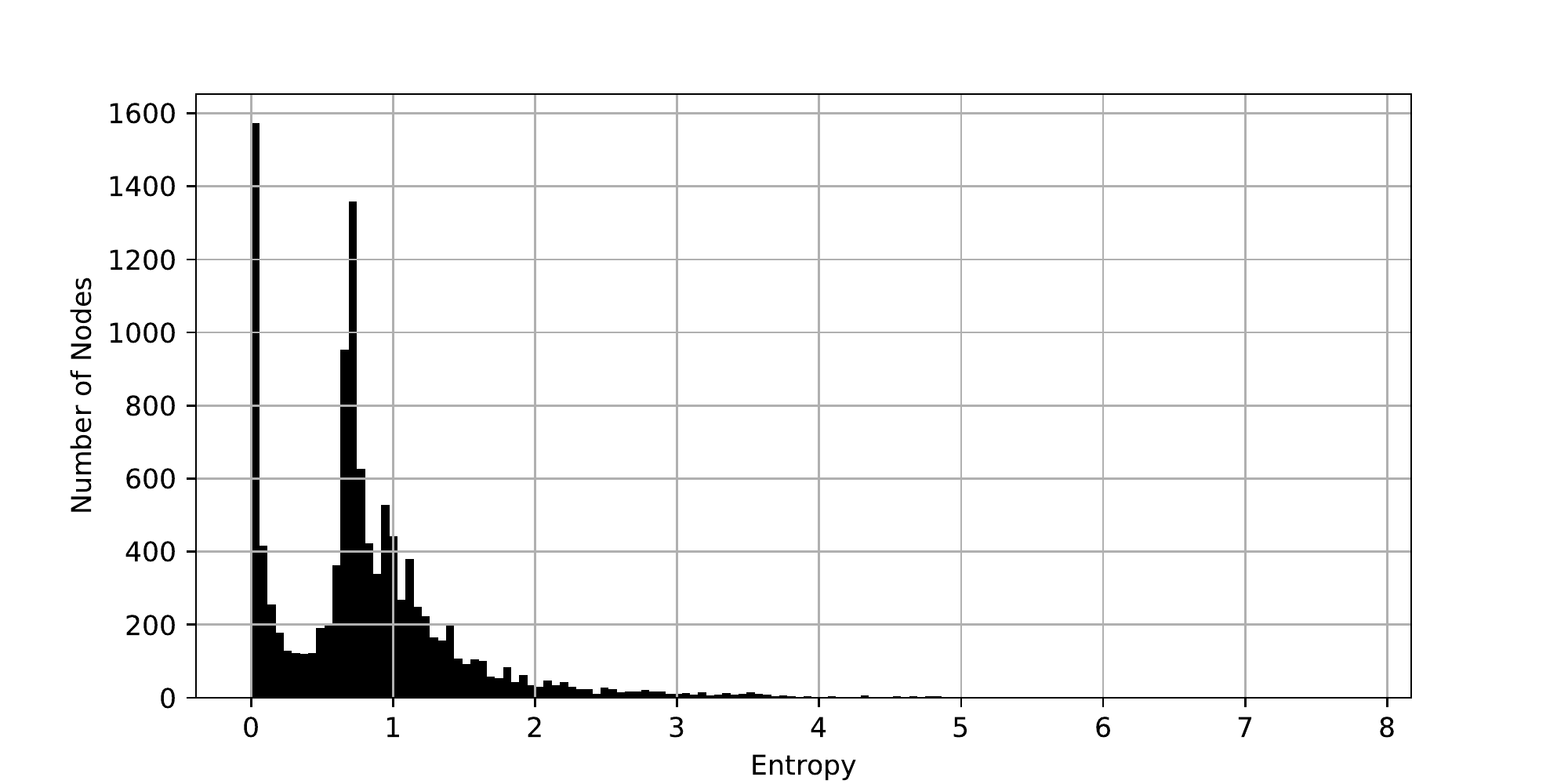}
		\caption{\system layer 1.}\label{fig:a}		
	\end{subfigure}
	\quad
	\begin{subfigure}{0.45\textwidth}
		\centering
		\includegraphics[width=1\textwidth]{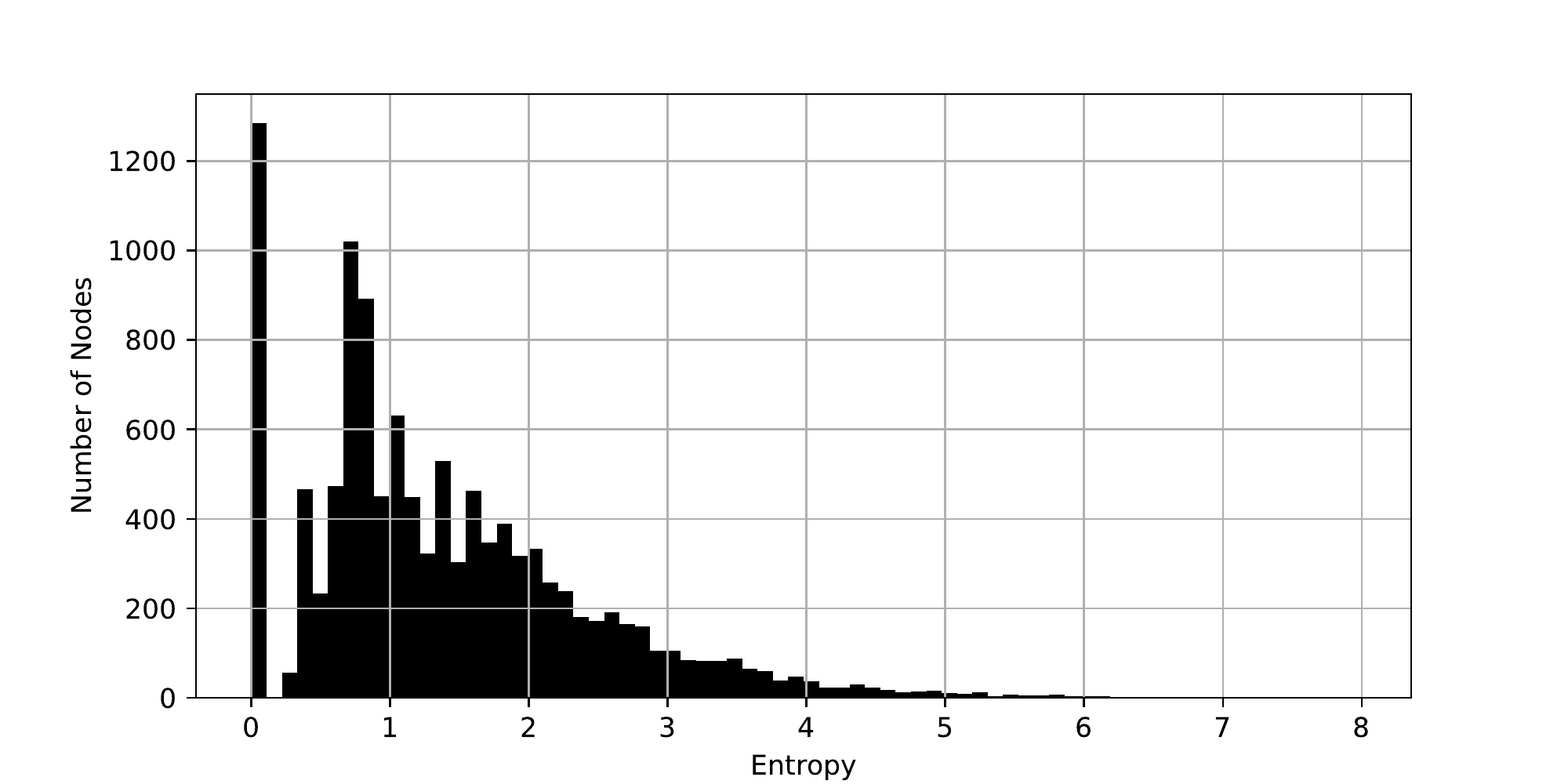}
		\caption{\system layer 2.}\label{fig:b}
	\end{subfigure}
    \caption{Entropy of attention distributions.}
    \label{fig:entropy}
\end{figure}

\subsection{Ablation Study}

We also conduct experiments to demonstrate the effect of each designed component on the model performance.


\noindent{\bf Attention Mechanism.} The \system network without attention mechanism is equivalent to R-GCN. To study the contribution of attention in learning, we use \system-ASA, which is a variant of \system trained with a random negative sampler, and compare it against R-GCN on Company dataset. As shown in Table~\ref{table:benchmarks}, \system-ASA outperforms R-GCN across all datasets in every measure proving that the attentions are effective. For further analysis, we measure the entropy of attention distribution over nodes. Figure~\ref{fig:entropy}(a,b) shows the node-wise attention entropy distribution learned by two \system layers. Note that we use a single head attention setting in this experiment. The plots have a distribution of lower entropy which means high degrees of concentration. Low entropy attention distributions provide more selective message passing capability than simple pooling operations such as mean pooling. It shows that our model \system is able to handle complex neighborhood by learning a few neighbors' messages which are more important than others. This validates our claim that not all neighbors are important in a multi-relational graph.


\noindent{\bf ASA Negative Sampling.} Next, we test the effect of ASA by replacing it with a random negative sampler. This refers to the comparison between \system and \system-ASA in Table~\ref{table:benchmarks}. \system outperforms \system-ASA in AUC and AP. It sometimes has lower F1 because the models are actually only better with a lower cutoff threshold. With the hard negative samples identified by ASA in this case, \system models act ``cautiously'' with the score distribution moving towards the lower end of the scale. We also extend this test to other representative models from each category which refers to the comparisons between R-GCN and R-GCN+ASA, and DistMult and DistMult+ASA. The results in Table~\ref{table:benchmarks} demonstrates the effectiveness of ASA component.





%% file: main.bbl
\begin{thebibliography}{29}
\providecommand{\natexlab}[1]{#1}
\providecommand{\url}[1]{\texttt{#1}}
\providecommand{\urlprefix}{URL }
\expandafter\ifx\csname urlstyle\endcsname\relax
  \providecommand{\doi}[1]{doi:\discretionary{}{}{}#1}\else
  \providecommand{\doi}{doi:\discretionary{}{}{}\begingroup
  \urlstyle{rm}\Url}\fi

\bibitem[{Battaglia et~al.(2018)Battaglia, Hamrick, Bapst, Sanchez{-}Gonzalez,
  Zambaldi, Malinowski, Tacchetti, Raposo, Santoro, Faulkner,
  G{\"{u}}l{\c{c}}ehre, Song, Ballard, Gilmer, Dahl, Vaswani, Allen, Nash,
  Langston, Dyer, Heess, Wierstra, Kohli, Botvinick, Vinyals, Li, and
  Pascanu}]{mpgoogle}
Battaglia, P.~W.; Hamrick, J.~B.; Bapst, V.; Sanchez{-}Gonzalez, A.; Zambaldi,
  V.~F.; Malinowski, M.; Tacchetti, A.; Raposo, D.; Santoro, A.; Faulkner, R.;
  G{\"{u}}l{\c{c}}ehre, {\c{C}}.; Song, H.~F.; Ballard, A.~J.; Gilmer, J.;
  Dahl, G.~E.; Vaswani, A.; Allen, K.~R.; Nash, C.; Langston, V.; Dyer, C.;
  Heess, N.; Wierstra, D.; Kohli, P.; Botvinick, M.; Vinyals, O.; Li, Y.; and
  Pascanu, R. 2018.
\newblock Relational inductive biases, deep learning, and graph networks.
\newblock \emph{CoRR} abs/1806.01261.

\bibitem[{Cai and Wang(2018)}]{kbgan}
Cai, L.; and Wang, W.~Y. 2018.
\newblock {KBGAN:} Adversarial Learning for Knowledge Graph Embeddings.
\newblock In \emph{Proceedings of the 2018 Conference of the North American
  Chapter of the Association for Computational Linguistics: Human Language
  Technologies}, 1470--1480. Association for Computational Linguistics.

\bibitem[{Cen et~al.(2019)Cen, Zou, Zhang, Yang, Zhou, and Tang}]{gatne}
Cen, Y.; Zou, X.; Zhang, J.; Yang, H.; Zhou, J.; and Tang, J. 2019.
\newblock Representation learning for attributed multiplex heterogeneous
  network.
\newblock In \emph{Proceedings of the 25th ACM SIGKDD International Conference
  on Knowledge Discovery \& Data Mining}, 1358--1368.

\bibitem[{Chen, Wu, and Zaki(2020{\natexlab{a}})}]{chen2020iterative}
Chen, Y.; Wu, L.; and Zaki, M. 2020{\natexlab{a}}.
\newblock Iterative Deep Graph Learning for Graph Neural Networks: Better and
  Robust Node Embeddings.
\newblock \emph{Advances in Neural Information Processing Systems} 33.

\bibitem[{Chen, Wu, and Zaki(2020{\natexlab{b}})}]{chen2019reinforcement}
Chen, Y.; Wu, L.; and Zaki, M.~J. 2020{\natexlab{b}}.
\newblock Reinforcement learning based graph-to-sequence model for natural
  question generation.
\newblock \emph{International Conference on Learning Representation} .

\bibitem[{Chen, Wu, and Zaki(2020{\natexlab{c}})}]{chen2020toward}
Chen, Y.; Wu, L.; and Zaki, M.~J. 2020{\natexlab{c}}.
\newblock Toward Subgraph Guided Knowledge Graph Question Generation with Graph
  Neural Networks.
\newblock \emph{arXiv preprint arXiv:2004.06015} .

\bibitem[{Dettmers et~al.(2018)Dettmers, Minervini, Stenetorp, and
  Riedel}]{conve}
Dettmers, T.; Minervini, P.; Stenetorp, P.; and Riedel, S. 2018.
\newblock Convolutional 2D Knowledge Graph Embeddings.
\newblock In McIlraith, S.~A.; and Weinberger, K.~Q., eds., \emph{Proceedings
  of the Thirty-Second {AAAI} Conference on Artificial Intelligence},
  1811--1818. {AAAI} Press.

\bibitem[{Dong, Chawla, and Swami(2017)}]{metapath2vec}
Dong, Y.; Chawla, N.~V.; and Swami, A. 2017.
\newblock metapath2vec: Scalable Representation Learning for Heterogeneous
  Networks.
\newblock In \emph{Proceedings of the 23rd {ACM} {SIGKDD} International
  Conference on Knowledge Discovery and Data Mining}, 135--144. {ACM}.

\bibitem[{Hamilton, Ying, and Leskovec(2017)}]{sage}
Hamilton, W.~L.; Ying, Z.; and Leskovec, J. 2017.
\newblock Inductive Representation Learning on Large Graphs.
\newblock In \emph{Advances in Neural Information Processing Systems},
  1024--1034.

\bibitem[{Hayashi and Shimbo(2017)}]{complex}
Hayashi, K.; and Shimbo, M. 2017.
\newblock On the Equivalence of Holographic and Complex Embeddings for Link
  Prediction.
\newblock In \emph{Proceedings of the 55th Annual Meeting of the Association
  for Computational Linguistics (Volume 2: Short Papers)}, 554--559.

\bibitem[{Hu et~al.(2020)Hu, Dong, Wang, and Sun}]{hgt}
Hu, Z.; Dong, Y.; Wang, K.; and Sun, Y. 2020.
\newblock Heterogeneous graph transformer.
\newblock In \emph{Proceedings of The Web Conference 2020}, 2704--2710.

\bibitem[{Ji et~al.(2020)Ji, Pan, Cambria, Marttinen, and Yu}]{kgsurvey}
Ji, S.; Pan, S.; Cambria, E.; Marttinen, P.; and Yu, P.~S. 2020.
\newblock A Survey on Knowledge Graphs: Representation, Acquisition and
  Applications.
\newblock \emph{CoRR} abs/2002.00388.
\newblock \urlprefix\url{https://arxiv.org/abs/2002.00388}.

\bibitem[{Kipf and Welling(2017)}]{gcn}
Kipf, T.~N.; and Welling, M. 2017.
\newblock Semi-Supervised Classification with Graph Convolutional Networks.
\newblock In \emph{5th International Conference on Learning Representations}.

\bibitem[{Kumar, Packer, and Koller(2010)}]{selfpaced}
Kumar, M.~P.; Packer, B.; and Koller, D. 2010.
\newblock Self-Paced Learning for Latent Variable Models.
\newblock In \emph{Advances in Neural Information Processing Systems},
  1189--1197.

\bibitem[{Li et~al.(2019)Li, Tao, Wu, Feng, Zhao, and Yan}]{adaptivens}
Li, J.; Tao, C.; Wu, W.; Feng, Y.; Zhao, D.; and Yan, R. 2019.
\newblock Sampling Matters! An Empirical Study of Negative Sampling Strategies
  for Learning of Matching Models in Retrieval-based Dialogue Systems.
\newblock In \emph{Proceedings of the 2019 Conference on Empirical Methods in
  Natural Language Processing and the 9th International Joint Conference on
  Natural Language Processing}, 1291--1296. Association for Computational
  Linguistics.

\bibitem[{Mikolov et~al.(2011)Mikolov, Kombrink, Burget, Cernock{\'{y}}, and
  Khudanpur}]{rnnlm}
Mikolov, T.; Kombrink, S.; Burget, L.; Cernock{\'{y}}, J.; and Khudanpur, S.
  2011.
\newblock Extensions of recurrent neural network language model.
\newblock In \emph{Proceedings of the {IEEE} International Conference on
  Acoustics, Speech, and Signal Processing}, 5528--5531. {IEEE}.

\bibitem[{Mikolov et~al.(2013)Mikolov, Sutskever, Chen, Corrado, and
  Dean}]{word2vec}
Mikolov, T.; Sutskever, I.; Chen, K.; Corrado, G.~S.; and Dean, J. 2013.
\newblock Distributed representations of words and phrases and their
  compositionality.
\newblock \emph{Advances in neural information processing systems} 26:
  3111--3119.

\bibitem[{Schlichtkrull et~al.(2018)Schlichtkrull, Kipf, Bloem, Van Den~Berg,
  Titov, and Welling}]{rgcn}
Schlichtkrull, M.; Kipf, T.~N.; Bloem, P.; Van Den~Berg, R.; Titov, I.; and
  Welling, M. 2018.
\newblock Modeling relational data with graph convolutional networks.
\newblock In \emph{European Semantic Web Conference}, 593--607. Springer.

\bibitem[{Sun et~al.(2019)Sun, Deng, Nie, and Tang}]{rotate}
Sun, Z.; Deng, Z.; Nie, J.; and Tang, J. 2019.
\newblock RotatE: Knowledge Graph Embedding by Relational Rotation in Complex
  Space.
\newblock In \emph{7th International Conference on Learning Representations}.
  OpenReview.net.

\bibitem[{Wang, Li, and Pan(2018)}]{igan}
Wang, P.; Li, S.; and Pan, R. 2018.
\newblock Incorporating {GAN} for Negative Sampling in Knowledge Representation
  Learning.
\newblock In \emph{Proceedings of the Thirty-Second {AAAI} Conference on
  Artificial Intelligence}, 2005--2012. {AAAI} Press.

\bibitem[{Wang et~al.(2019)Wang, Ji, Shi, Wang, Ye, Cui, and Yu}]{han}
Wang, X.; Ji, H.; Shi, C.; Wang, B.; Ye, Y.; Cui, P.; and Yu, P.~S. 2019.
\newblock Heterogeneous graph attention network.
\newblock In \emph{The World Wide Web Conference}, 2022--2032.

\bibitem[{Wu et~al.(2019)Wu, Pan, Chen, Long, Zhang, and Yu}]{gnnsurvey}
Wu, Z.; Pan, S.; Chen, F.; Long, G.; Zhang, C.; and Yu, P.~S. 2019.
\newblock A Comprehensive Survey on Graph Neural Networks.
\newblock \emph{CoRR} abs/1901.00596.
\newblock \urlprefix\url{http://arxiv.org/abs/1901.00596}.

\bibitem[{Xu et~al.(2019)Xu, Hu, Leskovec, and Jegelka}]{gin}
Xu, K.; Hu, W.; Leskovec, J.; and Jegelka, S. 2019.
\newblock How Powerful are Graph Neural Networks?
\newblock In \emph{7th International Conference on Learning Representations}.

\bibitem[{Yang et~al.(2015)Yang, Yih, He, Gao, and Deng}]{distmult}
Yang, B.; Yih, W.; He, X.; Gao, J.; and Deng, L. 2015.
\newblock Embedding Entities and Relations for Learning and Inference in
  Knowledge Bases.
\newblock In \emph{3rd International Conference on Learning Representations}.

\bibitem[{Yang et~al.(2020{\natexlab{a}})Yang, Xiao, Zhang, Sun, and
  Han}]{hgraphsurvey}
Yang, C.; Xiao, Y.; Zhang, Y.; Sun, Y.; and Han, J. 2020{\natexlab{a}}.
\newblock Heterogeneous Network Representation Learning: Survey, Benchmark,
  Evaluation, and Beyond.
\newblock \emph{CoRR} abs/2004.00216.
\newblock \urlprefix\url{https://arxiv.org/abs/2004.00216}.

\bibitem[{Yang et~al.(2020{\natexlab{b}})Yang, Ding, Zhou, Yang, Zhou, and
  Tang}]{samplednce}
Yang, Z.; Ding, M.; Zhou, C.; Yang, H.; Zhou, J.; and Tang, J.
  2020{\natexlab{b}}.
\newblock Understanding Negative Sampling in Graph Representation Learning.
\newblock \emph{CoRR} abs/2005.09863.

\bibitem[{Ying et~al.(2018)Ying, He, Chen, Eksombatchai, Hamilton, and
  Leskovec}]{pinsage}
Ying, R.; He, R.; Chen, K.; Eksombatchai, P.; Hamilton, W.~L.; and Leskovec, J.
  2018.
\newblock Graph convolutional neural networks for web-scale recommender
  systems.
\newblock In \emph{Proceedings of the 24th ACM SIGKDD International Conference
  on Knowledge Discovery \& Data Mining}, 974--983.

\bibitem[{Zhang et~al.(2019{\natexlab{a}})Zhang, Song, Huang, Swami, and
  Chawla}]{hetgnn}
Zhang, C.; Song, D.; Huang, C.; Swami, A.; and Chawla, N.~V.
  2019{\natexlab{a}}.
\newblock Heterogeneous graph neural network.
\newblock In \emph{Proceedings of the 25th ACM SIGKDD International Conference
  on Knowledge Discovery \& Data Mining}, 793--803.

\bibitem[{Zhang et~al.(2019{\natexlab{b}})Zhang, Yao, Shao, and
  Chen}]{nscaching}
Zhang, Y.; Yao, Q.; Shao, Y.; and Chen, L. 2019{\natexlab{b}}.
\newblock NSCaching: simple and efficient negative sampling for knowledge graph
  embedding.
\newblock In \emph{2019 IEEE 35th International Conference on Data Engineering
  (ICDE)}, 614--625. IEEE.

\end{thebibliography}
